\documentclass[10pt,conference]{IEEEtran}

\usepackage{amsmath,amssymb,amsfonts}
\usepackage{algorithmic}
\usepackage{algorithm}
\usepackage{graphicx}
\usepackage{textcomp}
\usepackage{xcolor}
\usepackage{booktabs}
\usepackage{multirow}
\usepackage[caption=false,font=footnotesize]{subfig}
\usepackage{hyperref}
\usepackage{placeins}
\hypersetup{
    hidelinks,
    pdfauthor={},
    pdftitle={},
    pdfsubject={},
    pdfkeywords={},
    pdfcreator={LaTeX},
    pdfproducer={pdflatex}
}

\newcommand{\resulttablestyle}{%
  \footnotesize
  \setlength{\tabcolsep}{3pt}
  \renewcommand{\arraystretch}{1.05}
}

\newcommand{\flowadam}{FlowAdam}
\newcommand{\RR}{\mathbb{R}}

\newcommand{\norm}[1]{\left\|#1\right\|}

\begin{document}

\title{FlowAdam: Implicit Regularization via\\Geometry-Aware Soft Momentum Injection}

\author{
  \IEEEauthorblockN{Devender Singh}
  \IEEEauthorblockA{Department of Computer Science\\
  Memorial University of Newfoundland\\
  St.\ John's, NL, Canada\\
  devenders@mun.ca}
  \and
  \IEEEauthorblockN{Tarun Sheel}
  \IEEEauthorblockA{Department of Mathematics and Statistics\\
  Memorial University of Newfoundland\\
  St.\ John's, NL, Canada\\
  tksheel@mun.ca}
}

\maketitle

\begin{abstract}
Adaptive moment methods such as Adam use a diagonal, coordinate-wise preconditioner based on exponential moving averages of squared gradients. This diagonal scaling is coordinate-system dependent and can struggle with dense or rotated parameter couplings, including those in matrix factorization, tensor decomposition, and graph neural networks, because it treats each parameter independently. We introduce FlowAdam, a hybrid optimizer that augments Adam with continuous gradient-flow integration via an ordinary differential equation (ODE). When EMA-based statistics detect landscape difficulty, FlowAdam switches to clipped ODE integration. Our central contribution is Soft Momentum Injection, which blends ODE velocity with Adam's momentum during mode transitions. This prevents the training collapse observed with naive hybrid approaches. Across coupled optimization benchmarks, the ODE integration provides implicit regularization, reducing held-out error by 10-22\% on low-rank matrix/tensor recovery and 6\% on Jester (real-world collaborative filtering), also surpassing tuned Lion and AdaBelief, while matching Adam on well-conditioned workloads (CIFAR-10). MovieLens-100K confirms benefits arise specifically from coupled parameter interactions rather than bias estimation. Ablation studies show that soft injection is essential, as hard replacement reduces accuracy from 100\% to 82.5\%.
\end{abstract}


\section{Introduction}

Adam~\cite{kingma2014adam} has become the default optimizer for deep learning, combining momentum with per-parameter adaptive learning rates via the update $\theta_{t+1} = \theta_t - \alpha \cdot m_t / (\sqrt{v_t} + \epsilon)$, where $m_t$ and $v_t$ are exponential moving averages of the gradient and squared gradient, respectively.

While effective on many problems, Adam's per-parameter scaling uses a diagonal preconditioner that is coordinate-system dependent---treating parameters as if they can be optimized independently~\cite{wilson2017marginal}. We call a problem \emph{coupled} when gradients for one parameter block depend strongly on others (i.e., large off-diagonal Hessian blocks or rotated valleys), so coordinate-wise scaling alone is insufficient. Adam's diagonal assumption can fail when parameters are correlated (requiring coordinated updates), when the loss landscape is rotated relative to parameter axes, or when saddle points require coordinated multi-parameter moves for escape. A canonical example is matrix factorization $A \approx UV^\top$, where gradients with respect to $U$ depend on $V$ and vice versa. Adam treats $U$ and $V$ independently, struggling to capture their coupling.

We propose \flowadam{}, a hybrid optimizer that addresses this limitation through three mechanisms:
\begin{itemize}
\item Adaptive mode switching: EMA-based detection of plateaus and ill-conditioned curvature triggers transitions from Adam to ODE integration.
\item Clipped ODE gradient flow: Solves a continuous-time gradient-flow ODE (with elementwise clipping for stability) using adaptive Runge-Kutta methods.
\item Soft momentum injection: Blends ODE velocity with Adam's existing momentum when returning from ODE mode, preserving scale knowledge and preventing training collapse.
\end{itemize}

We evaluate \flowadam{} on benchmarks spanning diverse coupling mechanisms: matrix completion ($UV^\top$ factorization), robust matrix factorization (Huber loss), tensor completion (trilinear coupling), GNN link prediction (topological coupling), inverse kinematics (trigonometric coupling), and real-world recommender data (Jester, MovieLens-100K). On problems where coupled optimization is the limiting factor, \flowadam{} achieves 10--22\% error reduction and outperforms tuned alternatives (Lion, AdaBelief) on real-world collaborative filtering. MovieLens-100K serves as a mechanism check: under a residualized protocol isolating the UV interaction term, \flowadam{} matches Adam, confirming benefits arise specifically from navigating coupled geometry. The method matches Adam on well-conditioned workloads (CIFAR-10: 91.7$\pm$0.1\%), confirming safe drop-in use. Source code is available at \url{https://github.com/idevender/flowadam}.

\section{Related Work}

\subsection{Adaptive Optimization Methods}

Adam~\cite{kingma2014adam} and its variants (AdaGrad~\cite{duchi2011adaptive}, AdamW~\cite{loshchilov2017decoupled}) use per-parameter learning rate adaptation. AdamW introduced decoupled weight decay, applying regularization directly to parameters rather than through gradients. Recent work has analyzed Adam's implicit second-order properties~\cite{malladi2022sqrtfree}, showing that the square-root in Adam's update relates to diagonal Fisher approximations; our work addresses coupling limitations complementary to this perspective. Other recent optimizers---AdaBelief~\cite{zhuang2020adabelief}, RAdam~\cite{liu2020radam}, Lion~\cite{chen2023lion}, Shampoo~\cite{gupta2018shampoo}, Sophia~\cite{liu2024sophia}---also operate in a purely discrete-time regime with diagonal or block-diagonal preconditioning, which may struggle on problems with dense, rotated parameter coupling.

Second-order methods such as L-BFGS~\cite{liu1989limited}, AdaHessian~\cite{yao2021adahessian}, and K-FAC~\cite{martens2015optimizing} can capture parameter correlations but introduce significant memory/compute overhead (e.g., large per-layer factors, Hessian-vector products, or expensive inversions), making them less practical as drop-in replacements at scale. Our goal is to provide a first-order, $O(N)$-memory optimizer that captures coupling benefits without full Hessian access. Matrix completion has specialized solvers such as ALS~\cite{koren2009matrix}; FlowAdam offers a general-purpose alternative without problem-specific algorithmic design.

\subsection{Continuous-Time Perspectives}

Our approach builds upon the continuous-time view of optimization, where discrete algorithms are interpreted as discretizations of ODEs~\cite{su2016differential,wibisono2016variational}. Scieur et al.~\cite{scieur2017integration} explicitly frame gradient descent and accelerated methods as numerical integration schemes, while Dherin et al.~\cite{dherin2025learning} discuss training via ODE solvers and integration choices. Unlike Neural ODEs~\cite{chen2018neural}, which use ODE solvers for model architectures, \flowadam{} uses continuous gradient flow as a trajectory-based velocity proposal to navigate ill-conditioned couplings that discrete steps struggle to resolve. Critically, we are not simply replacing Euler with Runge-Kutta; our contribution is triggered switching with momentum state reconciliation and gradient clipping. These continuous-time analyses are typically theoretical and do not yield practical optimizers with adaptive mode switching---which is our contribution.

\subsection{Prior Hybrid Approaches}

\flowadam{} relates to Lookahead~\cite{zhang2019lookahead}, which interpolates between synchronized checkpoints. However, \flowadam{} blends continuous ODE velocity with discrete momentum using adaptive triggers. Prior attempts to combine global search (like ODEs) with local search often failed because the local optimizer's momentum buffer becomes invalid after a global step, causing training instability. Our Soft Momentum Injection addresses this gap by treating the ODE trajectory as a velocity vector that blends smoothly into the momentum buffer.

\section{Method}

\subsection{Clipped Descent ODE Formulation}

We model optimization as a dynamical system. Define elementwise gradient clipping as $[\mathrm{clip_{grad}}(g)]_i = \mathrm{sign}(g_i) \min\{|g_i|, 1\}$, which clamps each coordinate to $[-1, 1]$. The clipped descent ODE is:
\begin{equation}
\frac{d\theta}{dt} = -\mathrm{clip_{grad}}(\nabla L(\theta))
\label{eq:gradient_flow}
\end{equation}
Note this is not the true gradient flow due to clipping, but it still guarantees descent (proved in Proposition~1 below). Unlike discrete steps, ODE integration provides a more continuous trajectory through the local gradient vector field. All norms $\|\cdot\|$ denote the $L_2$ norm unless otherwise specified.

\subsection{Adaptive Mode Switching}

\flowadam{} maintains EMA statistics of gradient behavior:
\begin{align}
\bar{g}_t &= \beta_{ema} \cdot \bar{g}_{t-1} + (1-\beta_{ema}) \cdot \norm{\nabla L_t} \\
\bar{c}_t &= \beta_{ema} \cdot \bar{c}_{t-1} + (1-\beta_{ema}) \cdot \norm{\nabla L_t - \nabla L_{t-1}}
\end{align}
where $\nabla L_t \triangleq \nabla L(\theta_t)$ and $\beta_{ema} = 0.9$ is the EMA decay factor (distinct from Adam's $\beta_1, \beta_2$), $\bar{g}_t$ tracks average gradient norm, and $\bar{c}_t$ measures step-to-step gradient variation---a lightweight trigger proxy that combines signals from curvature, step size, and noise (direct curvature estimation would require Hessian-vector products, outside our first-order goal). ODE mode triggers when:
\begin{equation}
\underbrace{\norm{\nabla L_t} < \alpha_s \cdot \bar{g}_t}_{\text{Plateau detected}} \quad \lor \quad \underbrace{\norm{\nabla L_t - \nabla L_{t-1}} > \alpha_c \cdot \bar{c}_t}_{\text{High gradient variation}}
\end{equation}
where $\alpha_s$ (switch sensitivity) and $\alpha_c$ (variation sensitivity) are hyperparameters. In highly stochastic regimes, $\|\nabla L_t - \nabla L_{t-1}\|$ is noise-dominated, so we use this trigger primarily for low-noise/full-batch problems. This approach is scale-invariant, adaptive to problem-specific gradient scales, and noise-smoothed via EMA averaging. In highly stochastic regimes, variance-reduction techniques (e.g., SVRG-style gradient corrections) could further improve trigger reliability; we leave this exploration to future work.

\subsection{ODE Integration}

When triggered, we integrate~\eqref{eq:gradient_flow} using \texttt{dopri5}, a Dormand--Prince RK4/5 method:
\begin{equation}
\theta_{new} = \text{ODESolve}\left(\frac{d\theta}{dt} = -\mathrm{clip_{grad}}(\nabla L), \theta_{old}, [0, \alpha \cdot \tau]\right)
\end{equation}
where $\alpha$ is the learning rate and $\tau$ is the ODE time scale. We use tolerance $10^{-4}$ for both relative and absolute tolerances. The adaptive RK integration provides local error control and automatically reduces internal step size in rapidly changing regions. If integration fails (rare), we fall back to a standard Adam step.

\subsection{Soft Momentum Injection}

Naive hybrid approaches reset Adam's momentum after ODE steps, causing state mismatch---Adam loses learned gradient scale information and training collapses. Our solution blends ODE velocity with existing momentum:
\begin{equation}
m_t \gets (1-\gamma) \cdot m_t + \gamma \cdot \underbrace{\frac{\theta_{old} - \theta_{new}}{\alpha}}_{\text{ODE velocity}}
\label{eq:soft_injection}
\end{equation}
where $\gamma \in [0,1]$ is the injection weight on ODE velocity. In practice, we apply velocity clipping~\eqref{eq:clip_vel} before injection. Setting $\gamma=0$ recovers pure Adam momentum with no injection, while $\gamma=1$ uses full ODE velocity but may lose Adam stability. The default value $\gamma=0.5$ provides balanced blending. Note that $v_{ode}$ is $\tau$-scaled (since integration spans time $\alpha \cdot \tau$), so $\tau$ controls how strongly an ODE segment influences momentum. We do not update $v_t$ or the Adam step counter on ODE steps and apply velocity clipping to prevent ODE velocity from dominating during hand-off:
\begin{equation}
\mathrm{clip_{vel}}(v_{ode}) = \begin{cases}
v_{ode} & \text{if } \norm{v_{ode}} \leq 5 \cdot \norm{m_t} \\
v_{ode} \cdot \frac{5 \cdot \norm{m_t}}{\norm{v_{ode}}} & \text{otherwise}
\end{cases}
\label{eq:clip_vel}
\end{equation}
The clipping factor of 5 was selected via validation and is robust to factors in $[3, 10]$. This enables smooth transitions while preserving Adam's scale knowledge.

\textbf{Warmup.} We disable ODE triggering during an initial warmup period ($t \leq t_{warmup}$, default $t_{warmup}=10$). This serves two purposes. First, it allows Adam's momentum $m_t$ to accumulate meaningful gradient statistics before velocity clipping~\eqref{eq:clip_vel} is applied, since clipping relative to $\|m_t\|$ is uninformative when $m_t \approx 0$. Second, it avoids false ODE triggers before the EMA statistics $\bar{g}_t, \bar{c}_t$ have stabilized. The warmup period $t_{warmup} = 10$ is robust across $[5, 20]$ in our experiments. The complete procedure is summarized in Algorithm~\ref{alg:flowadam}.

\begin{algorithm}[t]
\caption{\flowadam{} Optimizer}
\label{alg:flowadam}
\begin{algorithmic}[1]
\REQUIRE Learning rate $\alpha$, EMA decay $\beta_{ema}$, sensitivities $\alpha_s, \alpha_c$, injection weight $\gamma$, ODE time scale $\tau$, warmup steps $t_{warmup}$
\STATE Initialize $m_0 = 0$, $v_0 = 0$, $\bar{g}_0 = 0$, $\bar{c}_0 = 0$, $g_{\text{prev}} = 0$
\FOR{$t = 1, 2, \ldots$}
    \STATE $g_t \leftarrow \nabla L(\theta_t)$
    \STATE Update EMAs: $\bar{g}_t$, $\bar{c}_t$
    \STATE $\text{plateau} \leftarrow \norm{g_t} < \alpha_s \cdot \bar{g}_t$
    \STATE $\text{grad\_change} \leftarrow \norm{g_t - g_{\text{prev}}} > \alpha_c \cdot \bar{c}_t$
    \IF{$(\text{plateau} \lor \text{grad\_change})$ \AND $t > t_{warmup}$}
        \STATE $\theta_{new} \leftarrow \text{ODESolve}(-\mathrm{clip_{grad}}(\nabla L), \theta_t, \alpha \cdot \tau)$
        \STATE $v_{ode} \leftarrow (\theta_t - \theta_{new}) / \alpha$
        \STATE $m_t \leftarrow (1-\gamma) m_t + \gamma \cdot \mathrm{clip_{vel}}(v_{ode})$ \COMMENT{Soft inj.}
        \STATE $\theta_{t+1} \leftarrow \theta_{new}$ \COMMENT{$v_t$, step counter unchanged}
    \ELSE
        \STATE Standard Adam update (increments per-parameter step counter)
    \ENDIF
    \STATE $g_{\text{prev}} \leftarrow g_t$
\ENDFOR
\end{algorithmic}
\end{algorithm}

\subsection{Hyperparameter Guidelines}

We define two operating modes. \textbf{Mode A} (neural networks, stochastic) uses $\alpha_s = 0.4$, $\alpha_c = 3.0$, $\tau = 2.0$ for conservative triggering. \textbf{Mode B} (scientific ML, deterministic) uses $\alpha_s = 0.9$, $\alpha_c = 0.1$, $\tau = 0.5$ for aggressive triggering. This parallels Adam's default $\beta_1, \beta_2$ values---sensible defaults with problem-specific tuning when needed. Automatic mode selection based on estimated gradient noise level is a natural extension.

\subsection{Theoretical Properties}

We provide two formal results justifying the core mechanisms of \flowadam{}.

\textbf{Proposition 1} (Monotonic Descent for Clipped Gradient Flow). Let $L \in C^1(\RR^d)$ and define elementwise clipping $C: \RR^d \to \RR^d$ by $[C(g)]_i = \mathrm{sign}(g_i) \min\{|g_i|, 1\}$. Consider the clipped gradient flow ODE~\eqref{eq:gradient_flow}. Then along any differentiable trajectory $\theta(t)$:
\begin{equation}
\frac{d}{dt} L(\theta(t)) = -\nabla L(\theta(t))^\top C(\nabla L(\theta(t))) \leq 0
\end{equation}

\emph{Proof.} For each coordinate $i$, we have $\partial_i L \cdot [C(\nabla L)]_i = |\partial_i L| \min\{|\partial_i L|, 1\} \geq 0$ since clipping preserves sign. Summing over coordinates yields $\nabla L^\top C(\nabla L) \geq 0$, hence $\frac{d}{dt} L(\theta(t)) = -\nabla L^\top C(\nabla L) \leq 0$. $\square$

Note that clipping changes the vector field and we do not claim steepest descent; however, near critical points where $\|\nabla L\|_\infty < 1$, clipping is inactive and standard gradient flow is recovered. The continuous-time flow is monotone; numerically, we observe non-increase up to solver tolerances in our experiments (rare violations may occur due to discretization).

\textbf{Lemma 2} (Soft Injection Bound). If the momentum update is a convex blend $m^+ = (1-\gamma) m + \gamma \tilde{v}$ with $\gamma \in [0,1]$, then:
\begin{equation}
\|m^+\| \leq (1-\gamma)\|m\| + \gamma\|\tilde{v}\| \leq \max\{\|m\|, \|\tilde{v}\|\}
\end{equation}

\emph{Proof.} Triangle inequality gives $\|m^+\| \leq (1-\gamma)\|m\| + \gamma\|\tilde{v}\|$. Since $(1-\gamma) + \gamma = 1$, this convex combination satisfies $(1-\gamma)\|m\| + \gamma\|\tilde{v}\| \leq \max\{\|m\|, \|\tilde{v}\|\}$. $\square$

This bounds the momentum magnitude after injection: if $\tilde{v}$ (the ODE velocity) is additionally clipped by construction~\eqref{eq:clip_vel}, then $\|m^+\|$ is bounded by the clipping threshold, preventing momentum explosion during mode transitions. These component-level guarantees do not constitute a convergence proof for the full hybrid switching system; formal convergence analysis of the alternating Adam--ODE dynamics remains future work.

\section{Experiments}

We evaluate \flowadam{} on benchmarks organized into three groups: (A) mechanism validation, (B) conditioning and robustness, and (C) core challenge tasks demonstrating performance on coupled parameter problems. All experiments use PyTorch with \texttt{torchdiffeq} for ODE integration and standard bias-corrected Adam/AdamW implementations. We compare against Adam (and AdamW where relevant); SGD+momentum is included as a sanity baseline where applicable, along with problem-specific baselines. Each experiment reports mean $\pm$ std over multiple seeds unless noted.

\textbf{Experimental Protocol.} We use 5 seeds for all experiments. All matrix/tensor completion and robust matrix factorization experiments use full-batch (deterministic) optimization. ``Improv.'' denotes relative improvement vs.\ Adam: for RMSE, $(\text{Adam} - \text{FlowAdam})/\text{Adam} \times 100\%$; for AUC, $(\text{FlowAdam} - \text{Adam})/\text{Adam} \times 100\%$. Wall-clock times include ODE solver overhead. We report total gradient evaluations for compute-fair comparison: for Adam, total grad evals = number of steps (one backward per step); for \flowadam{}, total grad evals = outer steps + ODE NFE (number of function evaluations), since each \texttt{ode\_func} call triggers one backward pass. With \texttt{dopri5} at tolerance $10^{-4}$, each ODE trigger typically uses 8--12 NFE.

To ensure the ODE solver integrates the regularized landscape, \flowadam{} applies regularization directly to the loss ($\lambda \|\theta\|^2$). When AdamW is included, we tune its weight decay across $\{10^{-5}, 10^{-3}, 10^{-2}\}$. To control for regularization implementation differences, we additionally report ``Adam (explicit L2)'', which uses the same loss-based regularization as \flowadam{}: $L = \text{MSE} + \lambda \cdot (\|U\|^2 + \|V\|^2)$ with $\lambda = 10^{-5}$. This isolates the optimizer effect from regularization implementation.

\textbf{Evaluation Philosophy.} Our experiments serve four purposes: (1)~\emph{Drop-in evaluation}---comparing default \flowadam{} against default Adam/AdamW to assess practical adoption without tuning (Tables~I--VI); (2)~\emph{Alternative strategies}---comparing tuned \flowadam{} against fundamentally different optimization approaches (Table~VII); (3)~\emph{Conditioning stress-tests}---verifying behavior under ill-conditioning and confirming no regression on well-conditioned tasks (Section~IV.B); (4)~\emph{Mechanism checks}---toy diagnostics/ablations (Section~IV.A) and residualized MovieLens (Section~IV.C.7).

\subsection{Mechanism Validation}

\subsubsection{Rosenbrock Function}

The Rosenbrock function $f(x,y) = (1-x)^2 + 100(y-x^2)^2$ has a curved valley testing optimizer navigation. Starting at $(-1.5, 1.5)$ over 500 steps, \flowadam{} (Mode B) achieves 12\% lower final loss than Adam with 50 ODE triggers concentrated at the valley turn.

\subsubsection{Two Spirals Classification}

Two interleaved spirals ($1200^\circ$ rotation) test neural network optimization. Using 1000 points with a 3-layer MLP (24 hidden units, Tanh activation) over 4000 steps (Mode A), \flowadam{} achieves 100\% accuracy compared to Adam's 100\% and SGD's 54.4\%. \flowadam{} matches Adam on this task where adaptive methods excel, confirming no regression on well-conditioned neural network problems. (SGD's poor performance reflects sensitivity to per-parameter adaptation.)

\subsubsection{Ablation: Soft vs.\ Hard Replacement}

\begin{figure}[t]
\centering
\includegraphics[width=0.9\columnwidth, trim=0pt 0pt 0pt 0pt, clip]{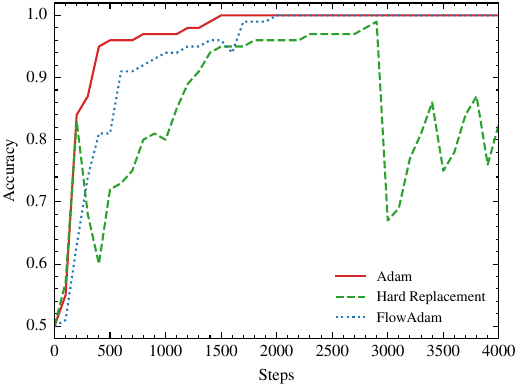}
\caption{Ablation: Hard replacement (green) induces a state mismatch by overwriting Adam's momentum buffer, causing a late-stage instability (final accuracy 82.5\%). \flowadam{} with soft injection (blue) blends the ODE velocity with the existing momentum, avoiding the mismatch and recovering stable convergence (100.0\%).}
\label{fig:ablation}
\end{figure}

Comparing soft momentum injection~\eqref{eq:soft_injection} against hard replacement ($m_t \leftarrow v_{\text{ode}}$) on two spirals reveals the importance of our approach. Hard replacement undergoes a late-stage collapse (down to 67\%) and ends at 82.5\%, while soft injection recovers to 100.0\% with fewer ODE triggers (221 for Hard vs.\ 166 for Soft). Figure~\ref{fig:ablation} illustrates this state mismatch problem and its resolution. Together with the compute-matched comparison in Section~IV.C.6 (Table~VI), this partially disentangles improvement sources: the compute-matched experiment isolates the ODE trajectory's contribution, while the present ablation isolates the injection mechanism.

\subsection{Conditioning and Robustness}

\subsubsection{Rotated Stiff Valley (Ill-Conditioned Quadratic)}

A 50-dimensional quadratic with random rotation and condition number 2000 directly tests Adam's diagonal assumption. We construct $L(\theta) = \frac{1}{2}\theta^\top H \theta$ where $H = Q \cdot \text{diag}(2000, 2000, 1, \ldots, 1) \cdot Q^\top$ and optimize over 500 steps (Mode B). \flowadam{} achieves final loss 36.0 compared to Adam's 97.0 and SGD's 156.2, yielding 2.7$\times$ lower loss than Adam. The high ODE trigger rate (96\%) indicates correct detection of the ill-conditioned landscape.

\subsubsection{CIFAR-10 Regression Test}

To verify \flowadam{} does not regress on well-conditioned problems, we evaluate on CIFAR-10 with ResNet-18 (50 epochs, batch 128, standard augmentation, no learning rate scheduling, Mode A). \flowadam{} achieves $91.7 \pm 0.1\%$ test accuracy compared to Adam's $91.8 \pm 0.1\%$ with negligible ODE triggers ($\approx 0$), confirming it is safe where Adam already works well. The near-zero trigger rate indicates that Mode~A's conservative thresholds effectively deactivate ODE integration on this well-conditioned stochastic task; extending the trigger to activate beneficially under mini-batch noise is an open direction.

\subsection{Coupled Parameter Benchmarks}

\subsubsection{Matrix Completion}
\label{sec:matrix}

Matrix completion---reconstructing a partially observed matrix via $A \approx UV^\top$---creates dense parameter coupling. We test three scenarios with increasing difficulty: Small Dense ($200 \times 300$, rank=10, 30\% observed), Medium Moderate ($300 \times 400$, rank=15, 20\% observed), and Larger Sparse ($400 \times 500$, rank=20, 15\% observed). We set model rank to true rank + 5 and train for 1000 full-batch steps. We evaluate RMSE on held-out entries (those not in the observation mask) using the known synthetic ground truth.

\begin{figure*}[t]
\centering
\includegraphics[width=0.85\textwidth]{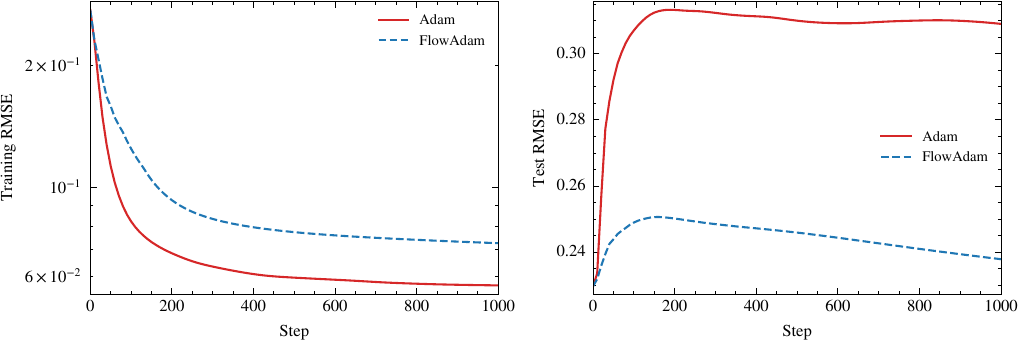}
\caption{Matrix Completion: Larger Sparse scenario (400$\times$500, rank=20, 15\% observed). Left: Training RMSE (log scale). Right: Test RMSE (linear scale). Representative single-seed run. Despite higher training RMSE (left), \flowadam{} achieves 23.0\% lower test RMSE (right; 5-seed mean: 21.8\%, Table~\ref{tab:matrix}), demonstrating implicit regularization.}
\label{fig:matrix}
\end{figure*}

\textbf{Results.} FlowAdam achieves 10.2--21.8\% lower test RMSE than Adam across all scenarios (Table~\ref{tab:matrix}), with improvement that scales with problem difficulty (Figure~\ref{fig:matrix}, illustrating the implicit regularization effect).

\textbf{Comparison with AdamW.} We swept AdamW weight decay $\lambda \in \{10^{-5}, 10^{-3}, 10^{-2}\}$ (Table~\ref{tab:matrix}). AdamW is insensitive to $\lambda$ (RMSE varies $<1\%$) and consistently underperforms Adam. A joint LR$\times\lambda$ sweep on Medium (30 configs) confirms this, with AdamW's best RMSE still 13\% worse than \flowadam{}.

AdamW's underperformance on matrix/tensor completion---yet parity with Adam on inverse kinematics (Table~\ref{tab:robotics})---suggests decoupled weight decay may interfere with implicit regularization dynamics specific to low-rank recovery, where the optimization trajectory itself induces bias toward simple solutions.

\textbf{Regularization control.} Adam with identical loss-based L2 regularization (explicit L2) achieves RMSE $0.116 \pm 0.001$ on Medium versus \flowadam{}'s $0.111 \pm 0.001$---a 4.3\% improvement confirming gains arise from ODE integration.

\begin{table}[t]
\centering
\caption{Matrix Completion: Test RMSE (5 seeds). Mode B.}
\label{tab:matrix}
\resulttablestyle
\begin{tabular}{@{}lcccc@{}}
\toprule
\textbf{Scenario} & \textbf{Adam} & \textbf{AdamW*} & \textbf{\flowadam{}} & \textbf{Improv.} \\
\midrule
Small Dense & $0.098\,{\pm}\,0.001$ & $0.115\,{\pm}\,0.003$ & $\mathbf{0.088\,{\pm}\,0.001}$ & 10.2\% \\
Medium Mod. & $0.129\,{\pm}\,0.001$ & $0.181\,{\pm}\,0.004$ & $\mathbf{0.111\,{\pm}\,0.001}$ & 14.0\% \\
Larger Sparse & $0.303\,{\pm}\,0.005$ & $0.640\,{\pm}\,0.015$ & $\mathbf{0.237\,{\pm}\,0.001}$ & 21.8\% \\
\bottomrule
\multicolumn{5}{@{}l@{}}{\footnotesize *Best AdamW across $\lambda \in \{10^{-5}, 10^{-3}, 10^{-2}\}$. Improv.\ vs.\ Adam.}
\end{tabular}
\end{table}

\subsubsection{Robust Matrix Factorization}
\label{sec:robust_pca}

Robust matrix factorization recovers a low-rank approximation $M \approx UV^\top$ from observations corrupted by heavy outliers. We test three scenarios with 20\% heavy outliers (magnitude 4--5$\times$ signal): Small ($60 \times 80$, rank=5), Medium ($80 \times 100$, rank=8), and Large ($100 \times 120$, rank=10). The loss uses Huber regression between observations and the low-rank reconstruction, which is robust to outliers.

\begin{table}[t]
\centering
\caption{Robust Matrix Factorization: Test RMSE (5 seeds). Mode B.}
\label{tab:robust_pca}
\resulttablestyle
\begin{tabular}{@{}lccc@{}}
\toprule
\textbf{Scenario} & \textbf{Adam} & \textbf{\flowadam{}} & \textbf{Improv.} \\
\midrule
Small Heavy & $0.736\,{\pm}\,0.053$ & $\mathbf{0.619\,{\pm}\,0.031}$ & 15.9\% \\
Medium Heavy & $0.955\,{\pm}\,0.020$ & $\mathbf{0.822\,{\pm}\,0.016}$ & 13.9\% \\
Large Heavy & $0.938\,{\pm}\,0.054$ & $\mathbf{0.819\,{\pm}\,0.014}$ & 12.7\% \\
\bottomrule
\end{tabular}
\end{table}

\flowadam{} achieves 12.7--15.9\% lower reconstruction error across all scenarios (Table~\ref{tab:robust_pca}). The dense coupling between $U$ and $V$ factor matrices creates correlated parameter updates where ODE integration excels, with $\sim$245 ODE triggers per run.

\subsubsection{Tensor Completion}
\label{sec:tensor}

Tensor completion extends matrix factorization to 3D: $\mathcal{T} \approx \sum_{r=1}^R u_r \circ v_r \circ w_r$, creating trilinear coupling across three factor matrices. We test Small Sparse ($30 \times 40 \times 50$, rank=5, 10\% observed), Medium Sparse ($40 \times 50 \times 60$, rank=8, 8\% observed), and Larger Sparse ($50 \times 60 \times 70$, rank=10, 8\% observed).

\begin{table}[t]
\centering
\caption{Tensor Completion: Test RMSE (5 seeds). Mode B.}
\label{tab:tensor}
\resulttablestyle
\begin{tabular}{@{}lcccc@{}}
\toprule
\textbf{Scenario} & \textbf{Adam} & \textbf{AdamW} & \textbf{\flowadam{}} & \textbf{Improv.} \\
\midrule
Small Sparse & $0.071\,{\pm}\,0.002$ & $0.083\,{\pm}\,0.008$ & $\mathbf{0.063\,{\pm}\,0.001}$ & 11.3\% \\
Medium Sparse & $0.068\,{\pm}\,0.002$ & $0.105\,{\pm}\,0.027$ & $\mathbf{0.060\,{\pm}\,0.001}$ & 11.8\% \\
Larger Sparse & $0.055\,{\pm}\,0.001$ & $0.067\,{\pm}\,0.005$ & $\mathbf{0.049\,{\pm}\,0.001}$ & 10.9\% \\
\bottomrule
\end{tabular}
\end{table}

\flowadam{} achieves 10.9--11.8\% improvement across all scenarios (Table~\ref{tab:tensor}). The 3D coupling amplifies correlated gradients beyond 2D matrices, with $\sim$498 ODE triggers demonstrating aggressive landscape exploration.

\subsubsection{GNN Link Prediction}
\label{sec:gnn}

Graph Neural Networks provide a test case where coupling arises from topological structure rather than matrix factorization. Using a 2-layer GCN with hidden dimension 32, we predict 20\% held-out edges on synthetic graphs. We test Medium Strong (500 nodes, degree=20), Large Moderate (600 nodes, degree=18), and Larger Challenge (800 nodes, degree=15).

\begin{table}[t]
\centering
\caption{GNN Link Prediction: Test AUC (5 seeds). Mode B.}
\label{tab:gnn}
\resulttablestyle
\begin{minipage}{0.85\columnwidth}
\centering
\begin{tabular}{@{}lccc@{}}
\toprule
\textbf{Scenario} & \textbf{Adam} & \textbf{\flowadam{}} & \textbf{Improv.} \\
\midrule
Medium Strong & $0.770\,{\pm}\,0.008$ & $\mathbf{0.793\,{\pm}\,0.009}$ & 3.0\% \\
Large Moderate & $0.730\,{\pm}\,0.009$ & $\mathbf{0.758\,{\pm}\,0.014}$ & 3.8\% \\
Larger Challenge & $0.698\,{\pm}\,0.013$ & $\mathbf{0.723\,{\pm}\,0.011}$ & 3.6\% \\
\midrule
\textit{Easy Calib.} & $0.941\,{\pm}\,0.003$ & $0.942\,{\pm}\,0.003$ & -- \\
\bottomrule
\end{tabular}\\[1mm]
{\footnotesize Calibration uses high-signal settings to verify the pipeline reaches high AUC.}
\end{minipage}
\end{table}

\flowadam{} achieves 3.0--3.8\% AUC improvement across the main scenarios (Table~\ref{tab:gnn}), demonstrating effectiveness beyond matrix-structured problems. The topological coupling from message passing creates correlated gradients that benefit from ODE integration (averaging 160--175 triggers per run). We use Mode~B (full-batch) on these moderate-sized synthetic graphs; for large-scale GNNs with mini-batch sampling, Mode~A would be appropriate. The calibration run confirms the model can reach high AUC ($>0.94$) when signal is abundant, ruling out capacity limitations.

\subsubsection{Inverse Kinematics}
\label{sec:robotics}

Multi-target inverse kinematics tests purely trigonometric coupling. For an $n$-link planar robot, end-effector position involves nested trigonometric functions---not bilinear $UV^\top$ structure. We optimize a trajectory of 10 waypoints for an 8-link arm with smoothness penalty $\lambda=1.0$ coupling all configurations (1500 steps).

\begin{table}[t]
\centering
\caption{Inverse Kinematics: Target RMSE (5 seeds). Mode B.}
\label{tab:robotics}
\resulttablestyle
\begin{minipage}{0.75\columnwidth}
\centering
\begin{tabular}{@{}lccc@{}}
\toprule
\textbf{Optimizer} & \textbf{Target RMSE} & \textbf{Median} & \textbf{Improv.} \\
\midrule
Adam & $0.182\,{\pm}\,0.080$ & $0.208$ & -- \\
AdamW & $0.182\,{\pm}\,0.080$ & $0.208$ & -- \\
\textbf{\flowadam{}} & $\mathbf{0.144\,{\pm}\,0.070}$ & $\mathbf{0.193}$ & $\mathbf{20.9\%}$ \\
\bottomrule
\end{tabular}\\[1mm]
\end{minipage}
\end{table}

\flowadam{} outperforms on 4/5 trajectory instances (Table~\ref{tab:robotics}), achieving 20.9\% mean improvement in target RMSE (7.2\% median). The trigonometric forward kinematics creates a highly non-convex landscape with many local minima; on one trajectory instance, \flowadam{} escapes a poor local minimum that traps all baselines (RMSE 0.016 vs.\ 0.192). This validates that benefits extend to geometric/trigonometric coupling.

\subsubsection{Jester: Real-World Matrix Completion}
\label{sec:jester}

Jester~\cite{goldberg2001eigentaste} is a classic collaborative filtering benchmark. We evaluate on Subset~1, which contains the most active users (24,983 users who rated 36+ jokes), resulting in 1.81 million ratings on 100 jokes (scale $[-10, 10]$). Unlike MovieLens-100K, Jester's small item count (100 jokes vs.\ 1,682 movies) reduces item bias variance, creating a setting where $UV^\top$ interaction learning dominates. We evaluate on held-out observed ratings, following standard practice for real collaborative filtering benchmarks without full ground truth.

\begin{table}[t]
\centering
\caption{Jester: Test RMSE (5 seeds). Mode B.}
\label{tab:jester}
\resulttablestyle
\begin{tabular}{@{}lcc@{}}
\toprule
\textbf{Optimizer} & \textbf{Test RMSE} & \textbf{Improv.} \\
\midrule
Adam (1K steps) & $4.731\,{\pm}\,0.014$ & -- \\
Adam Ext (4,350 grad evals) & $5.007\,{\pm}\,0.019$ & $-5.8\%$ \\
\textbf{\flowadam{}} (1K steps) & $\mathbf{4.427\,{\pm}\,0.009}$ & \textbf{6.4\%} \\
\bottomrule
\end{tabular}
\end{table}

\flowadam{} achieves 6.4\% lower RMSE than Adam (Table~\ref{tab:jester}), with improvement consistent across all 5 seeds (range 6.1--6.7\%). Notably, even when Adam's training is extended to match FlowAdam's compute budget (Adam Extended in Table VI), it fails to close the performance gap---instead overfitting rather than converging to a better solution. This indicates that \flowadam{}'s advantage stems from the implicit regularization of the ODE trajectory, not merely additional gradient evaluations. The compute-matched comparison shows \flowadam{} is 11.6\% better than Adam Extended with identical compute budget. The ODE triggering rate of 49.5\% (495/1000 steps) indicates a challenging landscape where coupled optimization matters.

\textbf{Comparison with alternative optimization strategies.} Tables~I--VI establish \flowadam{}'s advantage over Adam under the drop-in protocol. Table~\ref{tab:jester_modern} asks a different question. Does \flowadam{} remain competitive against methods employing fundamentally different optimization strategies? We compare against Lion~\cite{chen2023lion} (sign-based updates), AdaBelief~\cite{zhuang2020adabelief} (belief-scaled variance), and L-BFGS (quasi-Newton curvature approximation). \emph{Note: Different tuning protocols were used, so absolute RMSE values should not be compared across tables.}

\begin{table}[t]
\centering
\caption{Comparison with Alternative Optimization Strategies (Jester). All methods tuned via grid search (5 seeds).}
\label{tab:jester_modern}
\resulttablestyle
\begin{minipage}{0.75\columnwidth}
\centering
\resulttablestyle
\begin{tabular}{@{}lccc@{}}
\toprule
\textbf{Optimizer} & \textbf{Test RMSE} & \textbf{Std} & \textbf{vs.\ FlowAdam} \\
\midrule
Lion & $4.271$ & $0.008$ & $+2.0\%$ \\
AdaBelief & $4.495$ & $0.022$ & $+7.4\%$ \\
L-BFGS & $5.083$ & $0.017$ & $+21.4\%$ \\
\textbf{\flowadam{}} & $\mathbf{4.186}$ & $\mathbf{0.013}$ & -- \\
\bottomrule
\end{tabular}
\end{minipage}
\end{table}

Tuned \flowadam{} (4.19 RMSE) outperforms tuned Lion (4.27) by 2.0\% and tuned AdaBelief (4.49) by 7.4\%. L-BFGS performs poorly on this large-scale problem (1.81M ratings), likely due to difficulty approximating curvature with limited history vectors. Lion did not close the gap with extended training. \flowadam{} wins under \emph{both} protocols---untuned (6.4\% vs.~Adam) and tuned (2.0\% vs.~Lion)---demonstrating that gains arise from optimizer dynamics rather than Adam's limitations.

\begin{figure*}[!t]
\centering
\includegraphics[width=0.9\textwidth]{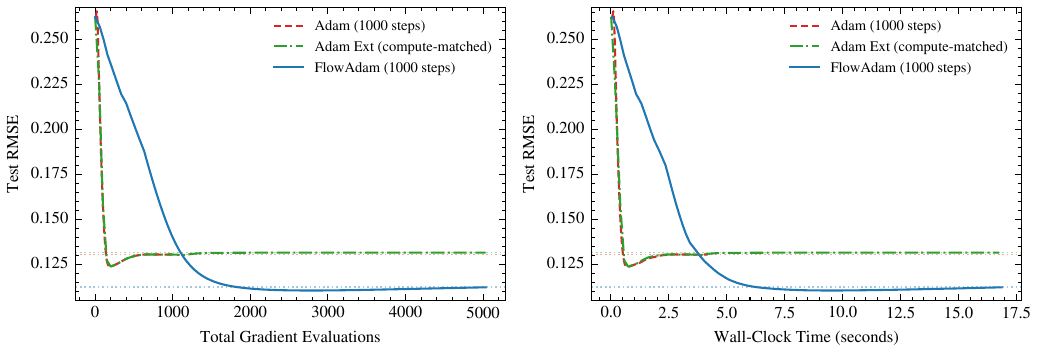}
\caption{Compute-matched comparison on Medium Matrix Completion. Left: RMSE vs.\ gradient evaluations. Right: RMSE vs.\ wall-clock time. Horizontal dotted lines indicate converged test RMSE.}
\label{fig:wallclock}
\end{figure*}

\subsubsection{MovieLens-100K (Residualized)}
\label{sec:movielens}

MovieLens-100K~\cite{harper2015movielens} serves as a \emph{mechanism check} rather than a headline gain. This dataset contains 100,000 ratings from 943 users on 1,682 movies. To directly test whether \flowadam{} helps coupled UV optimization, we employ a \textbf{residualized protocol}. First, we fit user/item biases ($\mu + b_u + b_i$) using Adam with early stopping. Then we compute residuals $e = r - (\mu + b_u + b_i)$. Finally, we train only the interaction term $UV^\top$ on residual targets.

MovieLens-100K (residualized protocol, Mode B, 5 seeds) shows that bias-only RMSE is $0.945 \pm 0.004$ using Adam. On residual UV targets, \flowadam{} matches Adam ($0.949 \pm 0.005$ vs.\ $0.949 \pm 0.005$), indicating no regression on a near-separable setting. This is \emph{consistent with our hypothesis}. \flowadam{}'s benefit is largest when coupled optimization is the limiting factor. On MovieLens, user/item biases explain most variance, leaving the residual UV component near the noise floor with little room for improvement.

\section{Discussion}

\subsection{Applicability and Scope}

\flowadam{} benefits full-batch or large-batch training with dense parameter correlations (e.g., matrix factorization) and ill-conditioned curvature, while providing no benefit (but no harm) when training is highly stochastic or Adam already converges well. For drop-in evaluation, \flowadam{} and Adam use identical default learning rates ($\alpha = 0.001$), ensuring performance differences reflect optimizer dynamics. \flowadam{} is intended as a drop-in, $O(N)$-memory replacement for Adam rather than a competitor to full second-order methods.

\textbf{Practitioner guidelines.} Use Mode~B for full-batch or large-batch training with low gradient noise, matrix/tensor factorization, and problems with dense off-diagonal Hessian structure. Use Mode~A for standard stochastic neural network training. Avoid when mini-batch noise would dominate ODE triggers.

\subsection{Computational Overhead}

ODE integration adds 8--12 gradient evaluations per trigger. On CIFAR-10 (Mode~A), triggers $\approx 0$ (no overhead). On matrix completion, $\sim$500 triggers provide implicit regularization. To verify gains are not from extra compute, we compared \flowadam{} (1,000 steps, 5,032 grad evals) against Adam Extended (5,032 steps) on Medium Matrix Completion. \flowadam{} achieves 14.7\% lower RMSE (Figure~\ref{fig:wallclock}), demonstrating the advantage comes from accessing better optima, not additional compute.

\subsection{Implicit Regularization}

The observation that frequent ODE triggers improve generalization despite higher training loss aligns with theoretical findings on implicit regularization~\cite{gunasekar2017implicit,arora2019implicit}. Gradient descent trajectories in matrix factorization induce implicit bias toward low-rank solutions. \flowadam{} enriches these dynamics by integrating continuous gradient flow, exploring coupled parameter geometry more effectively than discrete steps.

\subsection{Sensitivity Analysis}

The switch sensitivity $\alpha_s$ controls ODE frequency. Performance is stable for $\alpha_s \in [0.675, 0.90]$ with $<$1\% RMSE variation. However, $\alpha_s > 1$ causes plateau detection on nearly every step, collapsing the hybrid into always-ODE mode with degraded performance. We recommend $\alpha_s \leq 1$.

\begin{figure*}[!t]
\centering
\includegraphics[width=0.9\textwidth]{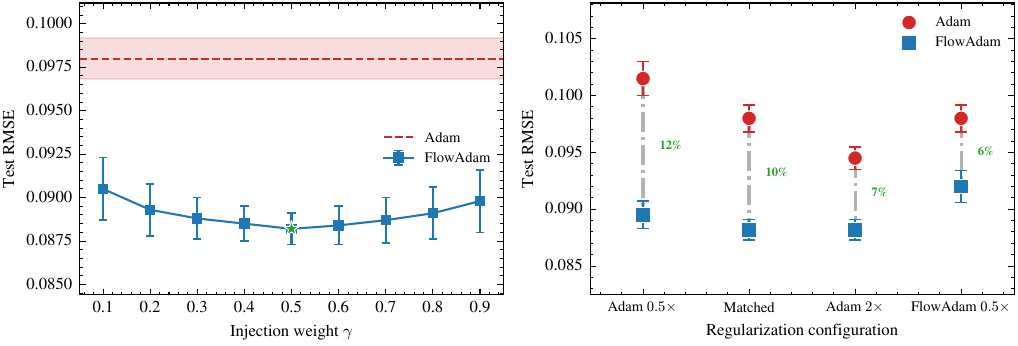}
\caption{Sensitivity analysis on matrix completion (Mode B, 5 seeds). Left: Performance is robust across $\gamma \in [0.1, 0.9]$; asterisk marks default $\gamma=0.5$. Right: \flowadam{} (fixed $\lambda=10^{-5}$) vs.\ Adam across regularization strengths. The x-axis shows regularization configurations: Adam at $0.5\times$, $1\times$ (matched), and $2\times$ the regularization strength of \flowadam{}, plus \flowadam{} at $0.5\times$ regularization. \flowadam{} outperforms Adam at all tested configurations.}
\label{fig:sensitivity}
\end{figure*}

For the injection weight $\gamma$ (weight on ODE velocity during momentum blending), all values in $[0.1, 0.9]$ achieve test RMSE within $\sim$2\% of the optimal (Figure~\ref{fig:sensitivity}, left), demonstrating that $\gamma=0.5$ is a robust default. Additionally, we verified robustness across regularization configurations. FlowAdam retained a $\sim$7\% advantage even when Adam was given $2\times$ regularization (Figure~\ref{fig:sensitivity}, right).

\subsection{Limitations}

Several limitations warrant discussion. First, the EMA-based triggering mechanism is sensitive to mini-batch noise: the gradient variation criterion $\|\nabla L_t - \nabla L_{t-1}\|$ conflates true curvature signals with stochastic variance, restricting reliable ODE triggering to low-noise or full-batch settings. Second, the hybrid switching system lacks formal convergence guarantees; our theoretical results (Proposition~1, Lemma~2) bound individual components but do not prove convergence of the alternating Adam--ODE dynamics. Third, ODE integration adds 8--12 gradient evaluations per trigger, which may be prohibitive for very large-scale architectures (e.g., Transformers) unless the trigger rate remains low. Fourth, users must manually select between Mode~A and Mode~B; automatic mode adaptation based on runtime noise estimation is a natural but unimplemented extension. Finally, our evaluation is limited to small- and medium-scale benchmarks; scaling behavior on modern large-scale training remains to be established.

\section{Conclusion}

We presented \flowadam{}, a hybrid optimizer combining Adam with ODE-based gradient flow. The key contribution, soft momentum injection, enables smooth mode transitions without destabilizing training. Across benchmarks spanning algebraic coupling (matrix/tensor factorization), architectural coupling (GNN), and geometric coupling (robotics), \flowadam{} achieves 10--22\% error reduction on matrix/tensor recovery, 3--4\% on GNN, and 6\% on real-world collaborative filtering (surpassing tuned alternatives including Lion) when coupled optimization is limiting. MovieLens residualized protocol confirms benefits arise specifically from navigating coupled geometry.

A notable finding is that aggressive ODE triggering provides implicit regularization, improving generalization with minimal explicit weight decay. Future directions include convergence analysis, adaptive $\gamma$ scheduling, and distributed training. More broadly, FlowAdam demonstrates that incorporating continuous-time dynamics into discrete optimizers can complement recent geometry-aware approaches to optimization.

\bibliographystyle{IEEEtran}
\bibliography{references}

@inproceedings{kingma2014adam,
  author    = {Diederik P. Kingma and Jimmy Ba},
  title     = {{Adam}: A Method for Stochastic Optimization},
  booktitle = {3rd International Conference on Learning Representations ({ICLR})},
  year      = {2015},
}

@article{duchi2011adaptive,
  author  = {John Duchi and Elad Hazan and Yoram Singer},
  title   = {Adaptive Subgradient Methods for Online Learning and Stochastic Optimization},
  journal = {Journal of Machine Learning Research},
  volume  = {12},
  pages   = {2121--2159},
  year    = {2011}
}

@inproceedings{loshchilov2017decoupled,
  author    = {Ilya Loshchilov and Frank Hutter},
  title     = {Decoupled Weight Decay Regularization},
  booktitle = {International Conference on Learning Representations ({ICLR})},
  year      = {2019}
}

@article{liu1989limited,
  author    = {Dong C. Liu and Jorge Nocedal},
  title     = {On the Limited Memory {BFGS} Method for Large Scale Optimization},
  journal   = {Mathematical Programming},
  volume    = {45},
  number    = {1},
  pages     = {503--528},
  year      = {1989},
  publisher = {Springer}
}

@inproceedings{yao2021adahessian,
  author    = {Zhewei Yao and Amir Gholami and Sheng Shen and Mustafa Mustafa and Kurt Keutzer and Michael Mahoney},
  title     = {{AdaHessian}: An Adaptive Second Order Optimizer for Machine Learning},
  booktitle = {Proceedings of the AAAI Conference on Artificial Intelligence},
  volume    = {35},
  number    = {12},
  pages     = {10665--10673},
  year      = {2021}
}

@inproceedings{martens2015optimizing,
  author       = {James Martens and Roger Grosse},
  title        = {Optimizing Neural Networks with {Kronecker}-Factored Approximate Curvature},
  booktitle    = {International Conference on Machine Learning ({ICML})},
  pages        = {2408--2417},
  year         = {2015},
  organization = {PMLR}
}

@article{chen2018neural,
  author  = {Ricky T. Q. Chen and Yulia Rubanova and Jesse Bettencourt and David K. Duvenaud},
  title   = {Neural Ordinary Differential Equations},
  journal = {Advances in Neural Information Processing Systems ({NeurIPS})},
  volume  = {31},
  year    = {2018}
}

@article{koren2009matrix,
  author    = {Yehuda Koren and Robert Bell and Chris Volinsky},
  title     = {Matrix Factorization Techniques for Recommender Systems},
  journal   = {Computer},
  volume    = {42},
  number    = {8},
  pages     = {30--37},
  year      = {2009},
}

@article{harper2015movielens,
  author    = {F. Maxwell Harper and Joseph A. Konstan},
  title     = {The {MovieLens} Datasets: History and Context},
  journal   = {ACM Transactions on Interactive Intelligent Systems (TiiS)},
  volume    = {5},
  number    = {4},
  pages     = {1--19},
  year      = {2015},
  publisher = {ACM}
}

@article{goldberg2001eigentaste,
  author    = {Ken Goldberg and Theresa Roeder and Dhruv Gupta and Chris Perkins},
  title     = {{Eigentaste}: A Constant Time Collaborative Filtering Algorithm},
  journal   = {Information Retrieval},
  volume    = {4},
  number    = {2},
  pages     = {133--151},
  year      = {2001},
  publisher = {Springer}
}

@article{zhang2019lookahead,
  author  = {Michael R. Zhang and James Lucas and Geoffrey Hinton and Jimmy Ba},
  title   = {Lookahead Optimizer: $k$ Steps Forward, 1 Step Back},
  journal = {Advances in Neural Information Processing Systems ({NeurIPS})},
  volume  = {32},
  year    = {2019}
}

@article{su2016differential,
  author  = {Weijie Su and Stephen Boyd and Emmanuel J. Cand\`{e}s},
  title   = {A Differential Equation for Modeling {Nesterov}'s Accelerated Gradient Method: Theory and Insights},
  journal = {Journal of Machine Learning Research},
  volume  = {17},
  number  = {153},
  pages   = {1--43},
  year    = {2016}
}

@article{scieur2017integration,
  author  = {Damien Scieur and Vincent Roulet and Francis Bach and Alexandre d'Aspremont},
  title   = {Integration Methods and Optimization Algorithms},
  journal = {Advances in Neural Information Processing Systems ({NeurIPS})},
  volume  = {30},
  year    = {2017}
}

@article{dherin2025learning,
  author  = {Benoit Dherin and Michael Munn and Hanna Mazzawi and Michael Wunder and Sourabh Medapati and Javier Gonzalvo},
  title   = {Learning by Solving Differential Equations},
  journal = {arXiv preprint arXiv:2505.13397},
  year    = {2025}
}

@article{wibisono2016variational,
  author    = {Andre Wibisono and Ashia C. Wilson and Michael I. Jordan},
  title     = {A Variational Perspective on Accelerated Methods in Optimization},
  journal   = {Proceedings of the National Academy of Sciences},
  volume    = {113},
  number    = {47},
  pages     = {E7351--E7358},
  year      = {2016},
  publisher = {National Academy of Sciences}
}

@article{zhuang2020adabelief,
  author  = {Juntang Zhuang and Tommy Tang and Yifan Ding and Sekhar C. Tatikonda and Nicha Dvornek and Xenophon Papademetris and James Duncan},
  title   = {{AdaBelief} Optimizer: Adapting Stepsizes by the Belief in Observed Gradients},
  journal = {Advances in Neural Information Processing Systems ({NeurIPS})},
  volume  = {33},
  pages   = {18795--18806},
  year    = {2020}
}

@article{chen2023lion,
  author  = {Xiangning Chen and Chen Liang and Da Huang and Esteban Real and Kaiyuan Wang and Yao Liu and Hieu Pham and Xuanyi Dong and Thang Luong and Cho-Jui Hsieh and Yifeng Lu and Quoc V. Le},
  title   = {Symbolic Discovery of Optimization Algorithms},
  journal = {Advances in Neural Information Processing Systems ({NeurIPS})},
  volume  = {36},
  pages   = {49205--49233},
  year    = {2023}
}

@inproceedings{gupta2018shampoo,
  author       = {Vineet Gupta and Tomer Koren and Yoram Singer},
  title        = {{Shampoo}: Preconditioned Stochastic Tensor Optimization},
  booktitle    = {International Conference on Machine Learning ({ICML})},
  pages        = {1842--1850},
  year         = {2018},
  organization = {PMLR}
}

@inproceedings{liu2024sophia,
  author    = {Hong Liu and Zhiyuan Li and David Hall and Percy Liang and Tengyu Ma},
  title     = {{Sophia}: A Scalable Stochastic Second-Order Optimizer for Language Model Pre-training},
  booktitle = {International Conference on Learning Representations ({ICLR})},
  year      = {2024}
}

@article{gunasekar2017implicit,
  author  = {Suriya Gunasekar and Blake E. Woodworth and Srinadh Bhojanapalli and Behnam Neyshabur and Nati Srebro},
  title   = {Implicit Regularization in Matrix Factorization},
  journal = {Advances in Neural Information Processing Systems ({NeurIPS})},
  volume  = {30},
  year    = {2017}
}

@article{arora2019implicit,
  author  = {Sanjeev Arora and Nadav Cohen and Wei Hu and Yuping Luo},
  title   = {Implicit Regularization in Deep Matrix Factorization},
  journal = {Advances in Neural Information Processing Systems ({NeurIPS})},
  volume  = {32},
  year    = {2019}
}

@article{wilson2017marginal,
  author  = {Ashia C. Wilson and Rebecca Roelofs and Mitchell Stern and Nati Srebro and Benjamin Recht},
  title   = {The Marginal Value of Adaptive Gradient Methods in Machine Learning},
  journal = {Advances in Neural Information Processing Systems ({NeurIPS})},
  volume  = {30},
  year    = {2017}
}

@inproceedings{malladi2022sqrtfree,
  author    = {Sadhika Malladi and Kaifeng Lyu and Abhishek Panigrahi and Sanjeev Arora},
  title     = {On the {SDEs} and Scaling Rules for Adaptive Gradient Algorithms},
  booktitle = {Advances in Neural Information Processing Systems ({NeurIPS})},
  year      = {2022}
}

@inproceedings{liu2020radam,
  author    = {Liyuan Liu and Haoming Jiang and Pengcheng He and Weizhu Chen and Xiaodong Liu and Jianfeng Gao and Jiawei Han},
  title     = {On the Variance of the Adaptive Learning Rate and Beyond},
  booktitle = {Proceedings of the 8th International Conference on Learning Representations ({ICLR})},
  year      = {2020}
}

\end{document}